\documentclass[conference]{IEEEtran}
\IEEEoverridecommandlockouts
\usepackage{cite}
\usepackage{amsmath,amssymb,amsfonts}
\usepackage{algorithmic}
\usepackage{graphicx}
\usepackage{textcomp}
\usepackage{xcolor}
\usepackage{booktabs}
\usepackage{multirow}
\usepackage{hyperref}
\usepackage{CJK}
\usepackage{threeparttable}
\def\BibTeX{{\rm B\kern-.05em{\sc i\kern-.025em b}\kern-.08em
    T\kern-.1667em\lower.7ex\hbox{E}\kern-.125emX}}
\begin{document}

\title{An Uncoupled Training Architecture for \\
		Large Graph Learning
		\thanks{Work performed an internship at InplusLab in the School of Data and Computer Science of Sun Yat-sen University.}
	}

	\author{\IEEEauthorblockN{Dalong Yang$^1$,
				Chuan Chen$^1$, Youhao Zheng$^2$, Zibin Zheng$^1$, Shih-wei Liao$^3$}
		\IEEEauthorblockA{$^1$School of Data and Computer Science,
		Sun Yat-sen University\\
			$^2$Faculty of Engineering and IT, The University of Sydney\\
			$^3$Department of Computer Science and Information Engineering, National Taiwan University\\
ydlkevin@gmail.com;
chenchuan@mail.sysu.edu.cn;
yzhe1017@gmail.com;
zhzibin@mail.sysu.edu.cn;
liao@csie.ntu.edu.tw}}
	\maketitle

\maketitle

\begin{abstract}
	Graph Convolutional Network (GCN) has been widely used in graph learning tasks. However, GCN-based models (GCNs) is an inherently \textit{coupled training framework} repetitively conducting the complex neighboring aggregation, which leads to the limitation of flexibility in processing large-scale graph. With the depth of layers increases, the computational and memory cost of GCNs grow explosively due to the recursive neighborhood expansion. To tackle these issues, we present Node2Grids, a flexible \textit{uncoupled training framework} that leverages the independent mapped data for obtaining the embedding. Instead of directly processing the coupled nodes as GCNs, Node2Grids supports a more efficacious method in practice, mapping the coupled graph data into the independent grid-like data which can be fed into the efficient Convolutional Neural Network (CNN). This simple but valid strategy significantly saves memory and computational resource while achieving comparable results with the leading GCN-based models. Specifically, by ranking each node's influence through degree, Node2Grids selects the most influential first-order as well as second-order neighbors with central node fusion information to construct the grid-like data. For further improving the efficiency of downstream tasks, a simple CNN-based neural network is employed to capture the significant information from the mapped grid-like data. Moreover, the grid-level attention mechanism is implemented, which enables implicitly specifying the different weights for neighboring nodes with different influences. In addition to the typical transductive and inductive learning tasks, we also verify our framework on million-scale graphs to demonstrate the superiority of the proposed Node2Grids model against the state-of-the-art GCN-based approaches. 
\end{abstract}

\begin{IEEEkeywords}
Graph Convolutional Network, 
Uncoupled Training Framework,
Convolutional Neural Network,
Large-Scale Graph Learning
\end{IEEEkeywords}

\begin{figure*}[h]
	\centering
	\includegraphics[scale=0.570]{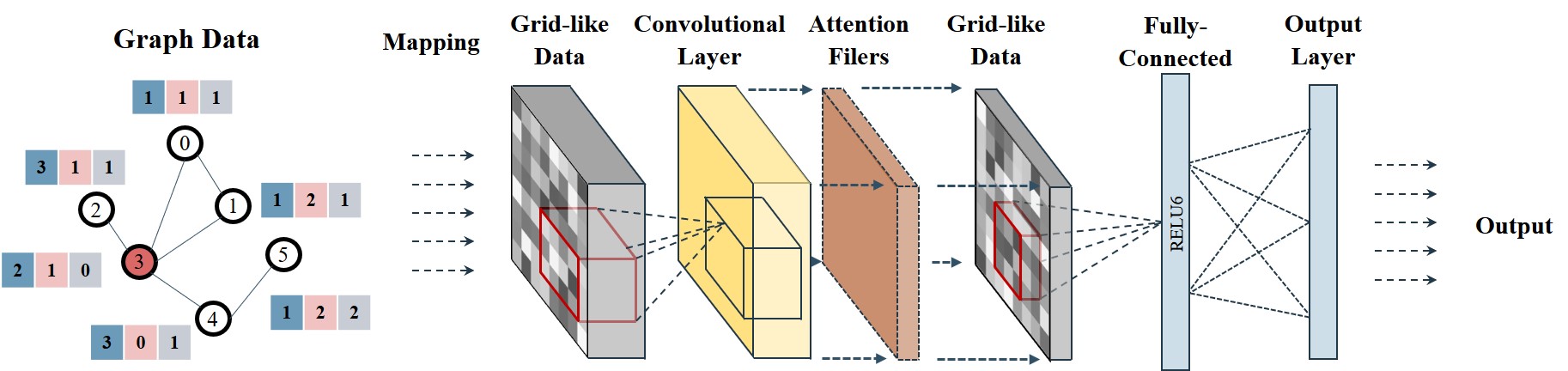}
	\caption{An illustration of the architecture for proposed Node2Grids. In this tiny example, there are 6 nodes which have 3 features. To begin with, the central node is mapped to the Euclidean structured grid with size of $k \times 1 \times 3$. Then a convolutional layer will be further used to extract the information from the gird-like data. In the next step, the proposed model employs an attention filters to learn the weight of each "pixel" (i.e. grid). Finally, the fully-connected layers are applied to gain an output for node classification.}
	\label{figure1}
\end{figure*}

\section{Introduction}
Graph Convolutional Network (GCN) \cite{b1} has achieved a great success in graph learning tasks, including node classification \cite{b26, b95}, link prediction \cite{b25, b96} and  community detection \cite{b97, b2}. By applying convolutional operations to gather the embeddings of neighbors layer by layer, GCN-based models (GCNs) significantly gain embeddings for the given nodes. However, GCNs are inherently coupled frameworks which repetitively propagate the representations through the interactions between neighbors during training, which leads to the challenges in practice due to the recursive neighborhood expansion. Specifically, unlike the calculations of conventional neural networks can be resolved into the uncoupled units (e.g., an image for CNN model), GCNs gain an embedding for a single node considering the repetitive aggregation of a great number of coupled nodes, leading to the inflexibility of training process. 

The original GCN is a full-batch framework, requiring the features and adjacency relations of nodes from the full graph Laplacian available, including the nodes for testing. And the matrix manipulations are operated over the whole graph. On one hand, with the layers increase, the receptive field size of GCN grows exponentially because of the recursive information aggregation from neighbors \cite{b98}, which leads to the greatly expensive requirements of computation and memory when processing large-scale graphs. On the other hand, full-batch GCN tends to leverage the fixed entire graph as a batch to train the model. However, for numerous cases, the graph is expanding dynamically rather than in a fixed state, which requires an inductive framework \cite{b70} capable of generalizing well to any augmentation graph by a significant model utilizing only training set for learning \cite{b9}.

Instead of taking the full-batch as inputs like original GCN, the mini-batch strategy \cite{b33} achieves better performance in flexibility when processing large-scale data and inductive learning tasks. Specifically, mini-batch training updates the learnable parameters only based on the sampled nodes in a mini-batch, which reduces the size of input data in each iteration to outperform in efficiency and memory requirement \cite{b5}. Inspired by these advantages, there are various studies attempting to introduce mini-batch strategy to GCN framework. GraphSAGE \cite{b4} proposes an inductive learning architecture and utilizes a fixed number of neighbors to aggregate feature information. FastGCN \cite{b9} samples a fixed number of nodes for each graph layer based on the node importance. LGCN \cite{b10} builds the sub-graphs for the training nodes by adjacent information, which reduces the batch size for the training progress. Recently, Cluster-GCN \cite{b5} obtains a great performance in large-scale inductive learning problems, sampling a block of nodes from the dense graph through the clustering algorithms. Even though the aforementioned approaches enable enhancing GCN to some extent, the flexibility of these GCN-based models are still restricted due to the essence of coupled framework, repeatedly considering the complex aggregations between neighboring nodes.  

Referring to mini-batch strategy, image processing tasks \cite{b37,b11} achieve good flexibility through various of advanced methods. When analyzing images, Convolotional Neural Networks \cite{b12} plays an important role to extract the meaningful information, whose local receptive field and sharing weights greatly reduce the number of parameters, which raise the learning efficiency. For taking advantage of image processing, numerous studies transform graph learning problems to "image" processing problems. Mathias Niepert et al. \cite{b13} propose to choose the sequence of fixed size nodes and build the significant neighborhood, using the CNN to extract the information. LGCN \cite{b10} constructs the grid-like data for nodes and applies CNN to learn the meaningful characters. DGCNN \cite{b14} utilizes conventional neural networks to process graph data after sorting graph vertices in a consistent order. Graph U-Nets \cite{b15} adopts the graph pooling approach to build an encoder-decoder model on graphs. hGANet \cite{b16} introduces the hard graph attention and channel-wise attention to overcome the limitation of consuming excessive computational resources. Yet such methods tend to continually aggregate information of neighbors by means of GCN,  which results in the inflexibility. In fact, if there exist an effective way of mapping the coupled nodes to an uncoupled "images", the calculation on each node for training process can only leverage the individual mapped data as the input of CNN, rather than aggregating the neighborhood embedding repetitively as GCN-based model. 

To overcome the defects of GCN-based models, we present Node2Grids, a flexible uncoupled architecture for large-scale graph learning. The framework of the proposed model consists of two parts: (1) mapping the coupled nodes (i.e. graph data) to independent grid-like data (i.e. Euclidean structured data), and (2) employing a simple three-layer neural network based on CNN to capture the characters from the grid-like data in Euclidean space. By the strategy of mapping, the classification loss on a single node is able to be resolved into the individual term on each sample, rather than depending on a number of coupled nodes like GCN-based models. Additionally, the downstream task enables applying mini-batch training method to the mapped grid-like data, which extends the proposed model to large-scale and inductive learning tasks.

Node2Grids constructs the corresponding independent grid-like data for each node, which greatly facilitates the efficiency of downstream training tasks. On one hand, Node2Grids realizes the extraction of neighboring information through the mapping step, avoiding the inflexibility of suffering recursive neighborhood expansion on training phase like GCN-based models. On the other hand, the Euclidean grids based training operation utilizes the uncoupled grid-like data for training, where the training nodes and testing nodes can be reconstructed respectively to keep the training and testing sessions separated from each other, which makes the model capable of predicting unseen nodes in the testing data (i.e. enables conducting inductive learning tasks). Furthermore, the batch size in the training nodes can be set elastically and the data in the same batch can be computed parallelly, which reduce the computational and memory requirements. In brief, the proposed Node2Grids not only adapts to transductive learning tasks but also have a strong ability to process large-scale graph and inductive learning tasks efficaciously.  

\begin{figure*}[ht]
	
	\centering
	
	\includegraphics[scale=0.5]{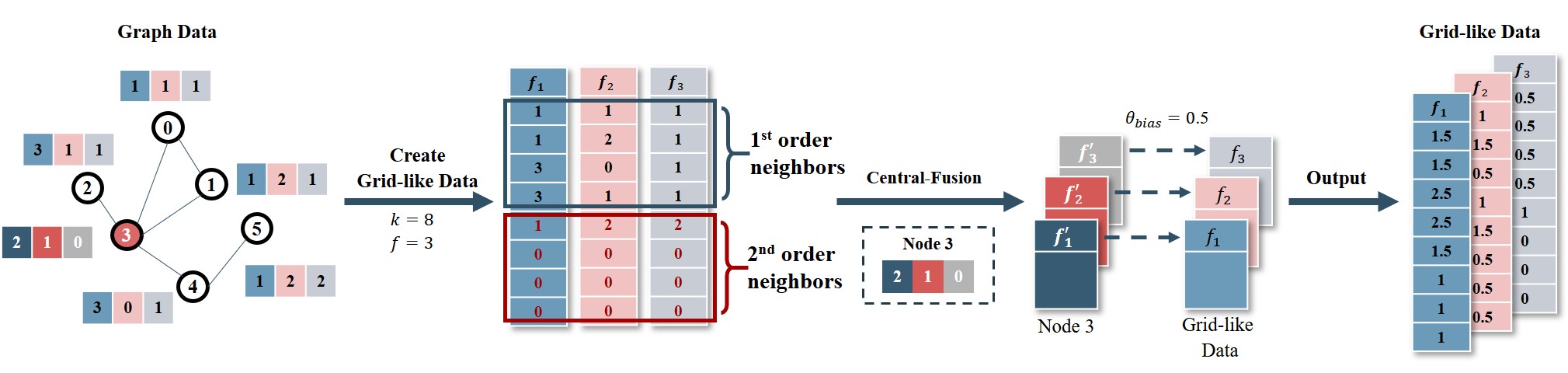}
	\caption{An example of the proposed mapping algorithm with $k=8$. The input graph has $6$ nodes with $3$ features (i.e. $f=3$), and the brown node 3 is the given central node. To begin with, Node2Grids  selects the first-order neighbors $N_1=\{0,1,2,4\}$ and second-order neighbors $N_2=\{5\}$. Then the $N_1$ and $N_2$ are ranked respectively according to the node degree. In grid-like data, each feature of the nodes represents a channel. For each channel, the proposed algorithm fills the feature of the nodes in $N_1$ and $N_2$ into the grids respectively to create the Euclidean structured data (pads the nodes from $N_2$ after padding the nodes from $N_1$). Note that if $k>\left|N_{1}\right|+\left|N_{2}\right|$, the default value of an unfilled grid is zero. In the last step, after creating the grid-like data with the size of $8 \times 1 \times 3$ , Node2Grids conducts information fusion (with $\theta_{bias} = 0.5$) between the central node and its neighbors, i.e. the features of the central nodes are introduced to update the grid-like data.}
	\label{figure2}
\end{figure*}

\begin{figure}[ht]
	\centering
	\includegraphics[scale=0.48]{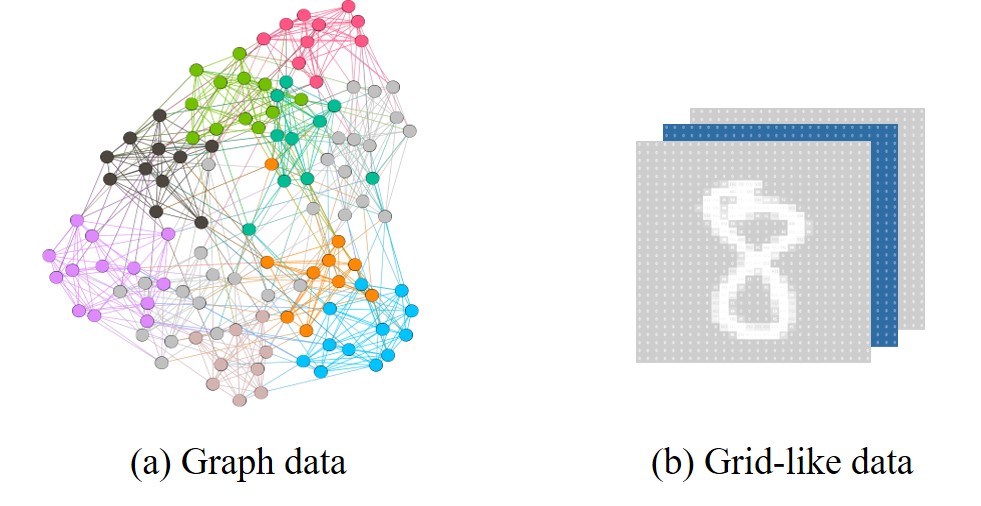}
	\caption{An example of real-word graph and grid-like data. The graph data is constructed by connected nodes. And the grid-like data consists of the regular grids.}
	\label{figure3}
\end{figure}

\section{Related Work}
Our model is related to previous graph learning techniques including Graph Convolutional Network, some recent developments in large-scale graph and inductive learning, and CNN-based frameworks. 

\textbf{Graph Convolutional Network.} The process of discrete convolution is essentially operations of weighted sum \cite{b19}. Based on the spectral domain \cite{b20}, GCN \cite{b1} introduces the convolutional operations on topological graph and reaches state-of-the-art performance on some datasets. Specifically, for a given graph, GCN gains a node representation by aggregating its neighbors' layer by layer. At each layer, the operations can be described as linear transformations and nonlinear activation functions. However, GCN is the coupled framework whose receptive field size grows exponentially as the layers increase, which causes a great limitation in efficiency and memory usage for large-scale graph learning. And the strategy of utilizing the fixed whole graph as a full-batch results in that original GCN is inapplicable for inductive learning tasks. 

\textbf{Large-scale graph and inductive learning.} Hamiltion et al. \cite{b4} propose the concept of inductive learning on large graph, where the nodes from testing set is unavailable during the training progress. In order to process inductive learning problems, GraphSAGE \cite{b4} trains the model through a fixed number of neighbors for each given node, where the representations are aggregated from the local neighborhood. At inference time, GraphSAGE employs the trained model to generate embeddings for entirely invisible testing nodes. While GraphSAGE still suffers from the high requirements of memory due to the exponential neighborhood growing problem. Graph Attention Network (GAT) \cite{b8} applies the attention mechanism to learn the attentions between central node and its neighbors. By gathering the features of neighbors layer by layer, GAT has an ability to specify the different weights for different neighboring nodes without costly matrix operations. However, just as GCN, the size of GAT's receptive field is upper-bounded by the network depth.

\textbf{CNN-based graph learning.} Convolutional Neural Network \cite{b12} is effective and efficient in analyzing Euclidean structured data. In order to introduce CNN to graph learning tasks, it is crucial to transform graph data to grid-like data. Mathias et al. \cite{b13} propose a framework to apply CNN on arbitrary graphs. They present a general method to transform graphs to locally connected regions. This framework applies the measures of centrality to labeling and ranking procedures, determining a strategy for mapping the graph data to vector representations by neighborhood assembly and graph normalization. But optimal graph normalization is a NP-hard problem which leads to the limitation in efficiency. LGCN \cite{b10} enhances the original GCN \cite{b1} in flexibility through the sub-graph selection strategy. In order to utilize CNN for extraction, LGCN selects a fixed number of neighboring nodes by ranking the feature values to construct the grid-like data for a given central node. Then, for LGCN, CNN is employed to obtain the significant information from the mapped grid-like data.  Even though LGCN leverages the useful strategies of sub-graphs building and CNN to enhance the model, there is still a drawback in flexibility due to the heavy pre-processing of implementing GCN to aggregate the neighboring characters. 

Instead of continual and complex neighboring aggregation as the GCN-based model (i.e., the coupled training framework), the proposed Node2Grids adopts a handy mapping approach to extract the characters of neighborhood (i.e., mapping the coupled nodes to uncoupled grid data). Moreover, Node2Grids applies the efficacious CNN-based network to extract the meaningful information from independent grid-like data. These strategies make the graph processing tasks efficient and flexible, achieving great performances in both transductive and inductive learning problems.

\section{Proposed Method}
In this section, we illustrate the framework of Node2Grids. The goal of Node2Grids is to construct a flexible framework to tackle the problem that conventional GCN-based models faced. The architecture of Node2Grids is shown in \autoref{figure1}. We first introduce the mapping strategy, and then describe the simple CNN-based three-layer neural network as well as the modified loss function we used for node classification tasks.

\subsection{Grid-like Data Generating}

In order to apply efficient CNN model (i.e. an uncoupled training method) to the coupled graph data, there are several challenges caused by the gaps between graph data and grid-like data. As shown in \autoref{figure3}, the graph is coupled with nodes and edges according to the topology, with various numbers of unordered neighborhood for each node. While the grid-like data comprise a fixed number of grids organized orderly and independently in space. Specifically, in generic graph, the number of each node’s neighbors is not constant, yet CNN requires the spatial neighborhood size remains the same. Besides, instead of the ordered neighboring grids in grid-like data, neighbors in graph cannot be sorted directly since there is no naturally recognized rule for node ranking. In this part, we illustrate the details for the mapping approach which transforms the graph data to uncoupled grid-like data. The algorithm of mapping can be categorized into three strategies, i.e. degree based selection, neighborhood expansion and Central-Fusion. Notably, the mapping processing can be conducted in parallel.

\textbf{Degree based selection.} We propose a degree-based method for mapping which transforms coupled graph data to independent Euclidean grids. As an inherent property of graph, the node degree reflects the influential ability \cite{b17} and the topology of original graph \cite{b80} naturally. And a node with higher degree tends to have a greater level of influence. Moreover, the node degree is a common character in graph data which makes the proposed model adaptive to various kinds of graph data.

\textbf{Neighborhood expansion.} Graphs are generally sparse in the natural world. We hold an opinion that the information included in the first-order neighbors is limited for sparse networks, which is not sufficient for expressing the character of a given node. Therefore, we introduce the second-order neighbors to extend the neighborhood for the given central node. To be specific, second-order neighbors are the nodes which can be reached from the given node by two hops, not included in the first-order neighbors and central node. In this case, we consider the secondary transmission of node characters to enrich the expression, which achieves obtaining the more expressive Euclidean grid.

\begin{table*}
	\caption{Summary of the datasets used in the experiments.}
	\label{table1}
	\setlength{\tabcolsep}{4mm}{
	\begin{tabular}{lccccccc}
		\toprule
		\textbf{Dataset}&\textbf{\#Nodes}&\textbf{\#Features}&\textbf{\#Classes}&\textbf{\#Training Nodes}&\textbf{\#Validation Nodes}&\textbf{\#Test Nodes}&\textbf{Task} \\
		\midrule
		Cora  & 2,708 & 1,433& 7& 140& 500& 1,000& Transductive \\
		\midrule
		Citeseer  & 3,327  & 3,703& 6& 120& 500& 1,000& Transductive \\
		\midrule
		Pubmed & 19,717 & 500& 3& 60& 500& 1,000 & Transductive \\
		\midrule
		PPI & 56,944  & 50& 121& 44,906 & 6,514& 5,524 & Inductive \\
		\midrule
		Amazon2M & 2,449,029 & 100 & 47 & 1,709,997 & N/A  & 739,032 & Inductive \\
		\bottomrule
		
	\end{tabular}}
\end{table*}

\textbf{Central-Fusion.} The proposed model constructs the grid-like data by leveraging the characters of central node and its neighborhood. Compared with the information carried by adjacent nodes, it is more important to consider the expression of the given central node. There should be two keys for fully representing the given node. First of all, the central node character should have more proportion than the neighboring nodes' in the mapped Euclidean structured data. Furthermore, the expression of central node needs to be contained on the global grid-like data rather than in local locations. Therefore, we introduce the approach of Central-Fusion to disperse the central node character into Euclidean grid globally. And we will demonstrate the effectiveness of Central-Fusion by extensive experiments in Section \uppercase\expandafter{\romannumeral4}-F and Section \uppercase\expandafter{\romannumeral4}-H. 

The proposed model builds a channel of grids for each feature. Supposing the given central node with $f$ features utilizes $k$ neighbors to construct the Euclidean structured grid, we set the dimension of the grid-like data to $k \times 1 \times f$, where $f$ is the number of grid channels (i.e. node features) and $k \times 1$ is the size of grid-like data (i.e. grid size) in each channel. The process of mapping is illustrated in \autoref{figure2}. For a given central node, we search for the first-order neighbors $N_1$ and second-order neighbors $N_2$ for padding, selecting the top $k$ neighbors from the searched nodes according to their degree value. Then the proposed Node2Grids pads Euclidean grid with features of the selected nodes.  Note that searching for second-order neighbors increases the time complexity, Node2Grids defines that the nodes in $N_1$ have the higher priority than $N_2$ when padding the grids, which enables  controlling whether to bring in second-order neighbors by adjusting the setup of $k$ according to the connectivity about network. Namely, if the graph is not sparse, the value of $k$ can be set in a proper range to only employ the first-order neighbors, which are sufficient to significantly express the original graph. Notably, if $k>|N_1|+|N_2|$, the default value of an unfilled grid is set to zero. We also conduct the exploration experiments about grid size (i.e. k) in Section \uppercase\expandafter{\romannumeral4}-F.

At the final step of mapping, the proposed Node2Grids conducts Central-Fusion, i.e. information fusion between the given node and the its neighbors. For the sake of expressing character of the given node around the grid-like data, we fuse current grids' features with the corresponding given node's. Concretely, 
\begin{equation}
G =\theta_{ {bias}} * G_{c}+\left(1-\theta_{ {bias}}\right) * G_{n},
\end{equation}
where $G_{n}\in\mathbb{R}^{k \times 1 \times f}$ is the grid-like data built by only selected neighbors. We obtain $G_{c} \in \mathbb{R}^{k \times 1 \times  f} $ extended from the central node, in which the grids of each channel are filled with the corresponding features of the central node. And $\theta_{ {bias }}$  is the bias coefficient of information fusion, which enable modifying the proportion of central node character.

\subsection{Grid-like Data Processing}
Convolutional Neural Network (CNN) is an efficiency and effective approach for Euclidean structured data processing. Inspired by these advantages, Node2Grids applies CNN to conduct  downstream tasks rather than GCN. At the training phase, we apply the 1-D convolutional kernels to extract the significant information from the uncoupled grid-like data. Specifically, Node2Grids employs a simple and flexible three-layer neural network, consisting of a convolutional layer and two fully-connected layers, to conduct the node classification tasks.  Moreover, we utilize the attention mechanism to learn the weights of each grid in grid-like data around all channels. 

\textbf{Grid-level attention mechanism.} For general Euclidean structured data such as images, "pixels" in different grids play different roles in expressing the "image" \cite{b100}. For example, the colored pixels of a image are generally more essential than the blank ones. In other words, there exist different biased weights for "pixels" in different space. Similarly, different neighbors have different influences to the central node \cite{b8}. Thus, it's significant to learn the grid-level attention (i.e. attentions on the neighbors with different level influences) for the mapped grid-like data in Node2Grids. To this end, we propose the attention mechanism which utilizes learnable attention filters to implicitly specify the weights within the grid-like data. Notably, all channels of the grid-like data share the same filter parameters due to the factor that Node2Grids focuses on attentions about neighbors rather than the features (the channels), which reduces the parameters greatly. 
Instead of prior frameworks' numerous attention operations on all neighboring nodes, a handy attention mechanism is implemented by Hadamard product on the structured grid-like data for Node2Grids. Moreover, we will prove the effectiveness of the grid-level attention mechanism by extensive experiments in Section \uppercase\expandafter{\romannumeral4}-G and Section \uppercase\expandafter{\romannumeral4}-H. 

\textbf{CNN-based network architecture.} After the process of mapping, the uncoupled Euclidean structured representations are obtained for nodes, where the general convolutional kernels can be applied to extract the advanced information. Compared with previous works applying GCN to aggregate the neighboring features, a simple CNN-based three-layer network architecture is employed to process the grid-like date, which enable speeding up the training progress and saving memory to enhance the scalability. To be specific, we apply a convolutional layer and two fully-connected layer in our architecture. Notably, the second fully-connected layers is used as the classifier. And the layer-wise propagation rule of Node2Grids is formulated as:  
\begin{equation}
\begin{split}
x_{l}&=\operatorname{Conv} (G),\\
{x}^{\prime}_{l}&=x_{l}+ (\frac{1}{h} \sum_{t=1}^{h}F_{a t t}) \circ x_{l}, \\
\tilde{x}_{l}&= g({x}^{\prime}_{l}),
\end{split}
\end{equation}
where $G \in \mathbb{R}^{k \times 1 \times f}$ is the mapped Euclidean structured data, in which $k \times 1$ represents the grid size in each channel and $f$ represents the number of channels. In this case, $G$ contains the fusion characters between the central node and its neighbors. $\operatorname{Conv}(\cdot)$ is a regular 1-D CNN that extracts the significant information from grid-like data, in which the kernel size is $n_{k e r }$  and the stride is set to 1,  without any padding strategy. Besides, Node2Grids employs the Hadamard product $\circ$ to realize attention mechanism. The $F_{a t t} \in \mathbb{R}^{(k-n_{ker}+1)\times 1\times f}$ is a single attention filter, where the number of learnable parameters is $(k-n_{ker}+1)\times1$ (i.e. all channels of grid-like date share the same filters' parameters). Note that the multi-head attention mechanism is applied, and the $h$ refers to the number of attention heads (i.e. the number of attention filters). In addition, $g(\cdot)$ is the function of fully-connected layers and $\tilde{x}_{l}$ is the output of the three-layer network.

\subsection{Modified Loss Function}
The learnable parameters in attention filters have a great impact on the output. In order to enhance the ability to resist disturbance, the greater coefficient is set for $l_2$ regularization in attention filters. For making the loss function adapted to attention mechanism, we modify it by multi-value $l_2$-regularization. The modified loss function of Node2Grids is defined as:
\begin{equation}
L_{total}=L(y, \bar{y})+\lambda \sum w^{2}+\lambda_{a t t} \sum w_{a t t}^{2},
\end{equation}
where $y$ and $\bar{y}$ are predicted labels and real labels of nodes respectively, $w$ and $w_{att}$ are the learnable parameters in neural networks and attention filters respectively. In addition, $\lambda$ and $\lambda_{att}$ are the $l_2$-regularization coefficients for the neural networks and attention filters respectively. $L(y, \bar{y})+\lambda \sum w^{2}$ is the general loss function with $l_2$-regularization. And $\lambda_{a t t} \sum w_{a t t}^{2}$ is the formula of $l_2$-regularization for attention filters to prevent over-fitting.  Notably, $\lambda_{{att}}$ is set to be larger than $\lambda$, which allows a stronger constraints to the parameters updating for attention filters.\\

\section{Experiments}
In this section, we evaluate the proposed Node2Grids in two classes of node classification tasks, i.e. transductive and inductive learning. We describe the datasets applied for node classification tasks and the setups of our model. In addition to demonstrate the effectiveness against the state-of-the-art GCN-based approaches, we also verify the high efficiency of the proposed model by comparing with leading approaches. Moreover, we further discuss the effect of hyper-parameters, the effectiveness of attention mechanism as well as the ablation studies.

\subsection{Datasets}

We conduct the experiments on both transductive and inductive learning datasets. The statistics of the datasets are summarized in \autoref{table1}. 

\textbf{Transductive learning dataset.} For transductive learning problems, nodes from the entire dataset are available during the training progress, including the nodes from testing or validation set whose labels are unseen.  We follow the recent studies and conduct the experiments over the three benchmark datasets, i.e. Cora, Citeseer and Pubmed datasets \cite{b90}. In these three citation networks, the nodes and edges represent documents and citations respectively. For the sake of ensuring the consistency, we follow the experimental settings of GCN \cite{b1} in the experiments, i.e., 20 nodes in each class are utilized for training, 500 nodes are utilized for validation and 1,000 nodes are utilized for testing.

\textbf{Inductive learning dataset.} In the inductive learning tasks, nodes from testing or validation set are unavailable for the training progress, which requires the model to trains a global function to obtain embeddings for testing or validation nodes. To this problem, we conduct the experiments on inductive graphs. Protein-protein interaction (PPI) dataset \cite{b91} contains graphs corresponding to different human tissues. The dataset consists of 20 training graphs, 2 validation graphs and 2 testing graphs. And there are multiple labels from 121 classes for each node. Moreover, we verify the scalability of the proposed framework on Amazon2M \cite{b5}, which is a million-scale network with more than 61 millions edges and 2 millions nodes. In the Amazon2M graph, the nodes  represent the products and the links represent the connected products are purchased together. Notably, the validation graphs and testing graphs are unseen during the training progress. 

\begin{table*}
\centering
	\caption{The Summary of results for transductive learning experiments in terms of average node classification accuracy, on the dateset of Cora, Citeseer and Pubmed. }
		\label{table2}
	\setlength{\tabcolsep}{10mm}{
		\begin{tabular}{lccc}
			\toprule
			\textbf{Method} &\textbf{Cora} & \textbf{Citeseer} & \textbf{Pubmed}\\
			\midrule
			DeepWalk \cite{b21}  & 67.2\% & 43.2\% & 65.3\% \\
			Chebyshev \cite{b22} &75.7\% & 64.7\% & 77.2\%  \\
			GCN \cite{b1} & 81.5\% & 70.3\% & 79.0\% \\
			GAT \cite{b8} & 83.0\% & 72.5\% & 79.0\% \\
			LGCN \cite{b10} & 83.3\% & 73.0\% & 79.5\% \\
			hGANet \cite{b16} & 83.5\% & 72.7\% & 79.2\% \\
			\midrule
			Node2Grids without Attenion & 83.1 $\pm$ 0.2\% & 72.4 $\pm$ 0.5\% & 76.3 $\pm$ 0.7\% \\
			\textbf{Node2Grids (Ours) } & \textbf{83.7 $\pm$ 0.2\%} & \textbf{73.2 $\pm$ 0.4\%} & \textbf{79.8 $\pm$ 0.5\%} \\
			\bottomrule
	\end{tabular}}
\end{table*}

\begin{table*}[]
	\centering
	\caption{Results of comparison in time cost and node classification accuracy with GCN and LGCN (using the sub-graph strategy for LGCN, i.e LGCN${sub}$), on the datasets of Cora, Citeseer and Pubmed.  The time cost is averaged from the start to the end of the training over 30 runs. Notably, the time cost consists of both mapping and training progress in the proposed Node2Grids model.}
	\linespread{1} 

	\renewcommand\arraystretch{1.3}
	\begin{tabular}{|c|ccc|ccc|ccc|ccc|}
		\cline{1-10}
		\multirow{1}{*}{} & 
		\multicolumn{3}{|c|}{\textbf{Cora}} &
		\multicolumn{3}{|c|}{\textbf{Citeseer}}&  
		\multicolumn{3}{|c|}{\textbf{Pubmed}} &  
		\\ \cline{1-10} 
		\multirow{2}{*}{GCN} & 
		\multicolumn{1}{|c|}{Batch size} & \multicolumn{1}{|c|}{Time} & Accuracy &
		\multicolumn{1}{|c|}{Batch size} & \multicolumn{1}{|c|}{Time} & Accuracy &
		\multicolumn{1}{|c|}{Batch size} & \multicolumn{1}{|c|}{Time}  & Accuracy & \\ \cline{2-10}
		& \multicolumn{1}{|c|}{2,708}&\multicolumn{1}{|c|}{\textbf{2.8s}}&\multicolumn{1}{|c|}{81.5\%}
		& \multicolumn{1}{|c|}{3,327}&\multicolumn{1}{|c|}{\textbf{4.6s}}&\multicolumn{1}{|c|}{70.3\%}
		& \multicolumn{1}{|c|}{19,717}&\multicolumn{1}{|c|}{20.9s}&\multicolumn{1}{|c|}{79.0\%}
		\\ \cline{1-10} 
		\multirow{2}{*}{LGCN$sub$} & 
		\multicolumn{1}{|c|}{Batch size} & 	\multicolumn{1}{|c|}{Time} & Accuracy &
		\multicolumn{1}{|c|}{Batch size} & 	\multicolumn{1}{|c|}{Time} & Accuracy &
		\multicolumn{1}{|c|}{Batch size} & 	\multicolumn{1}{|c|}{Time}  & Accuracy & \\ \cline{2-10}
		& \multicolumn{1}{|c|}{644}&\multicolumn{1}{|c|}{53.2s}&\multicolumn{1}{|c|}{83.3\%}
		& \multicolumn{1}{|c|}{442}&\multicolumn{1}{|c|}{27.6s}&\multicolumn{1}{|c|}{73.0\%}
		& \multicolumn{1}{|c|}{354}&\multicolumn{1}{|c|}{57.6s}&\multicolumn{1}{|c|}{79.5\%}
		\\ \cline{1-10} 
		\multirow{2}{*}{\textbf{Node2Grids (Ours) }} & 
		\multicolumn{1}{|c|}{Batch size} & 	\multicolumn{1}{|c|}{Time} & Accuracy &
		\multicolumn{1}{|c|}{Batch size} & 	\multicolumn{1}{|c|}{Time} & Accuracy &
		\multicolumn{1}{|c|}{Batch size} & 	\multicolumn{1}{|c|}{Time}  & Accuracy & \\ \cline{2-10}
		& \multicolumn{1}{|c|}{15}&\multicolumn{1}{|c|}{5.3s}&\multicolumn{1}{|c|}{\textbf{83.7 $\pm$ 0.2\%}}
		& \multicolumn{1}{|c|}{30}&\multicolumn{1}{|c|}{5.9s}&\multicolumn{1}{|c|}{\textbf{73.2 $\pm$ 0.4\%}}
		& \multicolumn{1}{|c|}{8}&\multicolumn{1}{|c|}{\textbf{7.3s}}&\multicolumn{1}{|c|}{\textbf{79.8 $\pm$ 0.5\%}}
		\\ \cline{1-10}

	\end{tabular}
	\label{table3}
\end{table*}

\begin{figure*}[ht]
	\centering
	\includegraphics[scale=0.52]{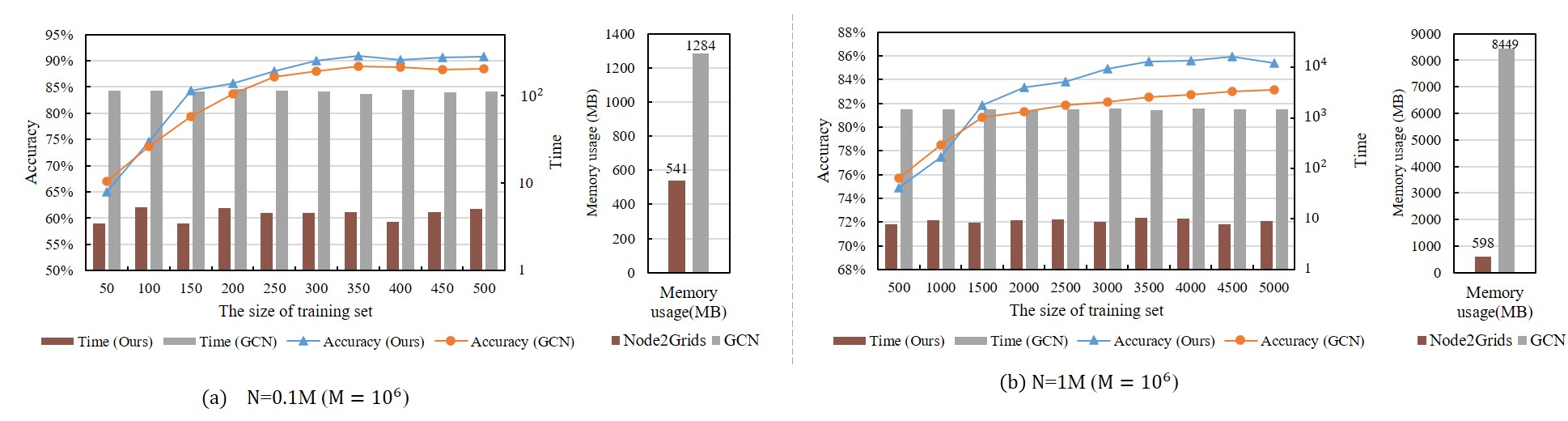}
	\caption{The results of experiments on large simulation networks, where N represents the size of dataset (i.e. the number of nodes in a network). The average node classification accuracy and time cost over 30 runs are reported in these experiments. In addition, the memory usages are reported on 0.1M and 1M graphs, with 500 and 5000 training nodes on these two graphs respectively.}
	\label{figure4}
\end{figure*}

\subsection{Baseline Methods}
DeepWalk \cite{b21} is a representative network embedding method, using the random walk to sample some nodes and obtains the embedding for the given node by these sampled nodes. Chebyshev \cite{b22}  and GCN \cite{b1} is the leading approaches for graph learning in spectral domain. Based on GCN, hGANet \cite{b16}  is the state-of-the-art model by introducing modified attention mechanism.  GraphSAGE \cite{b4}, GAT \cite{b8}  and LGCN \cite{b10} is the mini-batch based strategies, which can conduct the inductive learning tasks on large-scale graph. Notably, GraphSAGE-GCN \cite{b4} is an framework extending GCN to inductive learning tasks. And GraphSAGE-LSTM \cite{b4} is the GraphSAGE-based model employing LSTM as aggregators.

\subsection{Experimental Setup}

In this section, we describe the experimental setup in the proposed model in both transductive and inductive learning tasks, on the configuration of i7-7820X CPU and GTX 1080 Ti graphics card.

\textbf{Transductive learning task.} In stead of the full-batch based strategy, the proposed Node2Grids only leveraging partial nodes from the training set with batch size elastically set. For Cora, Citeseer and Pubmed datasets, the batch sizes are 15, 30 and 8 respectively. We apply RMSprop optimizer \cite{b92} to train the model, with the learning rate as 0.008 and weight decay as 0.0005. The dropout strategy \cite{b94} is applied in the first fully-connected layer, with the rate of 0.5. Note that the attention mechanism is introduced, there are 50 attention heads with the $\lambda_{att}$ for $l_2$ regularization in attention filters set to 0.0008, 0.025 and  0.07 for Cora, Citeseer and Pubmed datasets respectively.

\textbf{Inductive learning task.} We conduct the inductive learning experiments on protein-protein interaction (PPI) dataset and Amazon2M. The batch size is set to 2,000 and 80,000 in each iteration for PPI and Amazon2M datasets. The Nadam optimizer \cite{b93} is employed, in which the learning rate is set to 0.001 and the value of coefficient $\lambda$ is set to $5 \times 10^{-7}$ for $l_2$ regularization. In addition, The dropout \cite{b94} rate is set to 0.5 in both the convolutional layer and the first fully-connected layer. And there are also 50 attention heads applied with $\lambda_{att}$ set to $1 \times 10^{-6}$ for $l_2$ regularization in attention filters.

\subsection{Tranductive Learning tasks}

\subsubsection{Comparison of Effectiveness}
 In transductive learning problem, the results are reported by mean classification accuracy with the standard deviation over 100 runs as \cite{b1}. From the results summarized in \autoref{table2}, the proposed Node2Grids outperforms the current leading GCN-based models, especially achieving a great improvement over the original GCN by margins of 2.2\%, 2.9\% and 0.8\%  on the datasets of Cora, Citeseer and Pubmed respectively.
 
\subsubsection{Comparison of Computational Efficiency}
On transductive learning tasks, we also verify the high efficiency of the proposed model by comparison with previous GCN-based approaches. In addition to the three benchmark datasets (i.e., Cora, Citeseer and Pubmed datasets), we also conduct experiments on the large-scale simulation graphs. Notably, there are two factors having impacts on the time cost, which are (1) time spending for each epoch, (2) the convergence speed. To this end, we report the average time by the training progress from startup to obtain a valid testing model over 30 runs. For the sake of fairness, we take into account the time of generating the grid-like data in Node2Grids, and these three models do calculations on both training set (for training the model) and validation set (just for validation) in each epoch.

We implement the compared models by running the released codes from their github pages. The results of efficiency comparison on real-world networks are summarized in \autoref{table3}, where the LGCN$sub$ is LGCN model with the sub-graph generating algorithm to speedup the training progress.  From the results shown in \autoref{table3}, GCN achieves better efficiency on Cora and Citeseer datasets due to the parallel computations on small graphs. However, for Pubmed dataset, it's harder to process a large number of nodes for GCN, with efficiency greatly reduced. As for LGCN$sub$, it spends more time because of sampling a relatively large sub-graph for validation set which need to be processed by the coupled GCN framework. The performance  demonstrates the significant efficiency and effectiveness of the proposed Node2Grids model, especially in Pubmed which is a large graph with 19,717 nodes.

In order to further evaluate the scalability of Node2Grids, we conduct experiments on large-scale simulation networks. We generate nodes with labels based on the planted-l-partition model \cite{b50}, where the feature distributions are assigned on the basis of Madelon benchmark networks \cite{b51}. Specially, we refer to Pubmed dataset to set average node degree and class number to 6 and 3 respectively. Additionally, each node belongs to a single class with 500 features.  Two magnitudes of networks (with size of 0.1M and 1M) are employed during the experiments. And the compared model of GCN adopts two-layer convolutional networks in these simulation networks experiments. From the results reported in \autoref{figure4}, it can be observed that Node2Grids outperforms in efficiency with nearly 1/15 and 1/100 of the time cost in GCN on these two graphs respectively, while having greatly comparable performance in prediction accuracy. The memory costs are also reported in \autoref{figure4}, which takes into account the model parameters and all hidden representations for a batch. Compared with GCN, the memory usages of Node2Grids are only 1/14 of GCN on 1M network. Furthermore, both time cost and memory requirement for Node2Grids do not increase numerously when increasing the size of dataset, due to the strong parallel ability and few intermediate variables for the flexible training phase. Obviously, the gaps of memory usage as well as the time cost between these two models will further increase as the dataset expands. In conclusion, Node2Grids is a significantly low-overhead and highly parallel model, which demonstrates a superiority in processing large-scale transductive graph.

\subsection{Inductive Learning Tasks}
In inductive learning problem, we follow the study of \cite{b4} to report the micro-averaged F1 score (with standard deviation) of the nodes in unseen testing set over 10 runs. In addition to explore on the median size datasets, we also verify the scalability of the proposed model on the million-scale graph. 
\subsubsection{Median size
	dataset} We conduct the median size trial on PPI dataset.  From \autoref{table4}, it can be observed that the proposed Node2Grids performs better than previous methods, especially gaining a great progress over the representative GCN-based mini-batch approaches. (i.e., GraphSAGE-GCN and LGCN, by margins of 0.477 and 0.205 respectively).

\subsubsection{Million-scale dataset}  In order to further explore the performance of the proposed Node2Grids model on large inductive learning graph, we also conduct the experiments on Amazon2M network, which is the largest public dataset with more than 2 millions nodes. On this dataset, Node2Grids is compared with GraphSAGE-GCN, a representative GCN-based framework, in terms of effective, efficiency and memory cost, where the time and memory cost are calculated as the transductive learning tasks on Section IV-D. For Graphsage-GCN, we sample 9000 nodes and set the neighborhood size to 20 at most in each iteration due to the limitation of memory. Besides, we set the layer of GraphSAGE-GCN to one with the factor of the explosive growth of computational and memory resource with the layer increases. Note that the Amazon2M dataset doesn't partition the validation set, we randomly sample 5,000 nodes from the testing set as the validation data before the training. As we can see from \autoref{table5}, though the min-batch GCN-based framework of GraphSAGE-GCN enable conducting the inductive learning task, this coupled framework suffers the great computational and memory cost when processing large-scale graph due to the recursive neighborhood expansion. Compared with the coupled training framework, the time cost and memory usage of the proposed Node2Grids are respective only 1/2 and 3/10 of the GraphSAGE-GCN, meanwhile this uncoupled framework significantly outperforms the GraphSAGE-GCN on the aspect of effectiveness. Notably, with the layer increases, though  the higher F1 score of GraphSAGE-GCN may be obtained, the time and memory requirement will grow explosively which results in the resource constraints to apply the multi-layer GraphSAGE-GCN on such millon-scale graph.

\begin{table}
	\caption{The summary of results in terms of average micro-averaged F1 score for inductive learning tasks, on the dateset of PPI.}
	\label{table4}
	\centering
	\begin{tabular}{lc}
		\toprule
		\textbf{Method}& \textbf{PPI}\\
		\midrule
		GraphSAGE-GCN \cite{b4}& 0.500\\
		GraphSAGE-LSTM \cite{b4}  & 0.612\\
		LGCN \cite{b10} & 0.772\\
		GAT \cite{b8} & 0.973\\
		\midrule
		Node2Grids without Attention & 0.971 $\pm$ 0.003\\
		\textbf{Node2Grids (Ours)} & \textbf{0.977 $\pm$ 0.002}\\
		\bottomrule
	\end{tabular}
\end{table}

\begin{table}
	\caption{The summary of results in terms of average micro-averaged F1 score, time cost and memory usage for comparisons between GraphSAGE-GCN and Node2Grids, on the million-scale inductive dateset of Amazon2M.}
	\label{table5}
	\centering
	\begin{tabular}{cccc}
		\toprule
		\textbf{Method}&  \textbf{Time} & \textbf{Memory} & \textbf{F1 socre} \\
		\midrule
		GraphSAGE-GCN & 1,208s& 10,124MB& 0.766$\pm$0.005\\
		\midrule
		\textbf{Node2Grids (Ours)}  & \textbf{607s}& \textbf{3,121MB}& \textbf{0.878$\pm$0.003}\\
		\bottomrule
	\end{tabular}
\end{table}

\begin{figure*}[ht]

	\centering
	\includegraphics[scale=0.5]{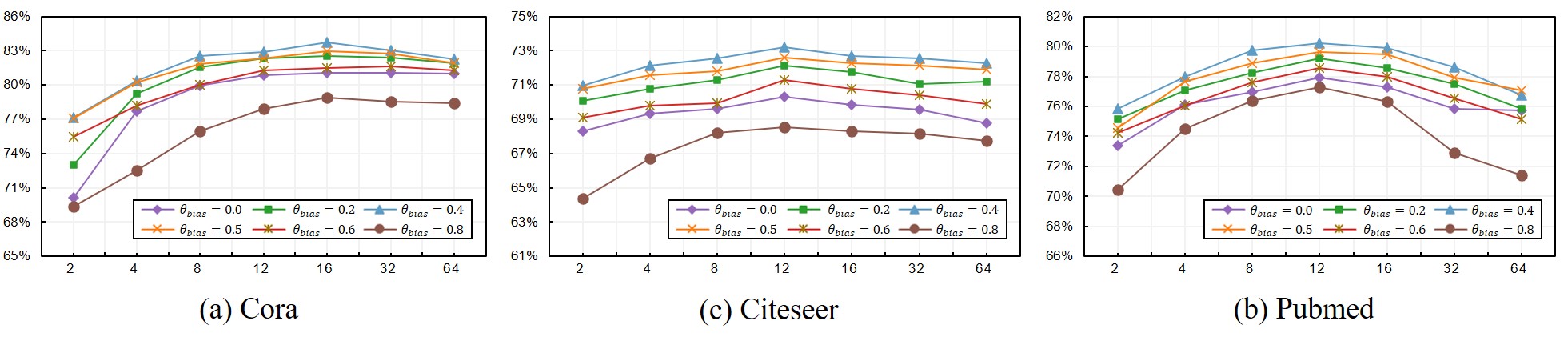}
	\caption{The results of exploring the influence of  $\theta_{bias}$ and $k$ on the datesets of Cora, Citeseer and Pubmed respectively. The setups described in Section  \uppercase\expandafter{\romannumeral4}-C are used. The node classification accuracy is reported in these experiments. The x-coordinate and y-coordinate represent $k$ and mean accuracy respectively. And the different colored lines represent the different occasions for $\theta_{bias}$. }
	\label{figure5}
\end{figure*}
\begin{table*}
	\caption{The results of ablation experoments. The mean node classification accuracy over 30 runs are reported on this table. The $\surd$ means taking the corresponding strategy, while the $\times$ is the opposite.}
	\label{table6}
	\centering
	\setlength{\tabcolsep}{4mm}{
		\begin{tabular}{cccccc}
			\toprule
			\textbf{\#Second-Order Neighbor}&\textbf{\#Central-Fusion}&\textbf{\#Attention Mechanism}&\textbf{Cora}&\textbf{Citeseer}&\textbf{Pubmed}
			\\
			\midrule
			$\times$  & $\times$ & $\times$ & 67.6\%& 63.8\%& 40.7\%\\
			$\surd$  & $\times$  & $\times$& 78.9\%& 69.5\%& 69.0\%\\
			$\times$ & $\surd$ & $\times$& 77.7\%& 64.1\%& 45.7\%  \\
			$\times$ & $\times$  & $\surd$& 73.2\%& 69.1\% & 66.1\% \\
			
			$\surd$ & $\surd$ & $\times$ & 83.1\% & 72.4\% & 76.3\%   \\
			$\surd$ & $\times$ & $\surd$ & 80.9\% & 70.3\% & 78.1\%   \\
			$\times$ & $\surd$ & $\surd$ & 78.2\% & 70.2\% & 76.0\%   \\
			$\surd$ & $\surd$ & $\surd$ & \textbf{83.7\%} & \textbf{73.2\%}  & \textbf{79.8\%}    \\
			\bottomrule
			
	\end{tabular}}
\end{table*}
\begin{figure}[ht]
	\centering
	\includegraphics[scale=0.48]{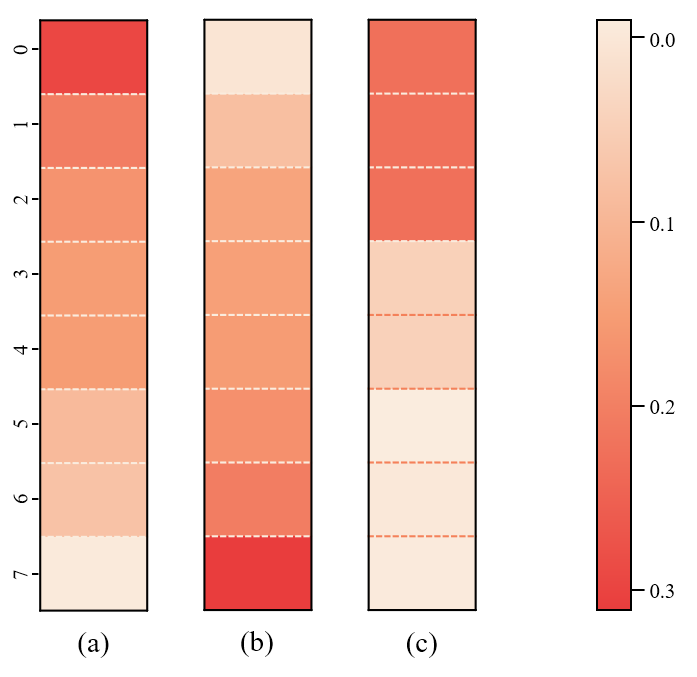}
	\caption{An example of the average convergent attention filters. We select the top 8 neighbors to build candidate sets which are ranked by node degree. The order we put the neighbors into grids is from the top to bottom. In (a) and (b), the nodes from candidate sets are putted to the grids in order and in reverse order respectively. The case of (c) is that the given central node features are put into the first 3 grids from the top to bottom and the other grids are filled with neighbors in order from the candidate set.}
	\label{figure6}
\end{figure}

\subsection{The Study of Central-Fusion and Grid Size}
In this section, we explore the influence of hyper-parameter $k$ and $\theta_{bias}$ on the datasets of Cora, Citeseer and Pubmed respectively. The Node2Grids utilizes central node fusion (Central-Fusion) to make sure that the character of the given central node can be expressed over the grid-like data globally, in which the $\theta_{bias}$ represents the level of central node fusion. (i.e. the greater $\theta_{bias}$, the more partition of central node character in the Euclidean grid). Besides, We select $k$ neighbors based on the value of degree and construct $k$-D grid-like data in each channel for the given node. Intuitively, the hyper-parameter $k$ represents the level of introducing neighbors’ information. We report node classification accuracy to evaluate the effect of $k$ and $\theta_{bias}$. The results are summarized in \autoref{figure5}.

For the effect of Central-Fusion, it can be obviously discovered that this strategy achieves positive influence by comparing the situation of $\theta_{bias}=0$ to a significant value of $\theta_{bias}$. From  \autoref{figure5}, the model performs best on these three datasets when $\theta_{bias} = 0.4$. Particularly, Node2Grids performs worse when the $\theta_{bias}$ is greater or less than the applicable value, which indicates that it exists a reasonable proportion for central node fusion. In other words, deficiency or redundant of the central node character will result in the deviation of expressing the raw graph. As for inductive learning tasks on PPI and Amazon2M datasets, there also exist an applicable setup for $\theta_{bias}$ with a value of 0.55 and 0.5 respectively.

On Cora, Citeseer and Pubmed datasets, these three networks are sparse with the average node degrees of 4, 5 and 6 respectively. It can be seen from the  \autoref{figure5} that $k$ is set appropriately with values of 16, 12 and 12 for Cora, Citeseer and Pubmed respectively. Notably, these best setups for $k$ bring in proper second-neighboring characters to enrich the grid-like data. At a word, the proposed Node2Grids model requires a reasonable grid size (i.e. $k$) to build the grid-like data. When $k$ is too small, it's difficult for the inadequate neighbors to reflect the origin graph. While the $k$ is too large, there are two aspects of decreasing the performance for Node2Grids: on one hand, the large $k$ means selecting various of second-order neighbors for mapping, resulting in introducing the redundant second-order information which causes a negative impact on expression of the closer first-order neighbors; On the other hand, the model may pads too much zero in the grid-like data, which compromises the performance of subsequent extraction task. Moreover, we also conduct the experiments to explore the setting of $k$ for PPI and Amazon2M datasets. Note that the best setup for k is 16 on both PPI and Amazon2M graphs which is smaller than their average node degrees (with the values of 31 and 25 respectively), indicating that the Euclidean grid can be built without introducing the second-order neighbors if the network is not sparse.

\subsection{The Study of Attention Mechanism}
During the phase of grid-like data processing, we employ the grid-level attention mechanism to learn the weights of grids in the grid-like data. As shown in \autoref{table2}, Node2Grids with attention mechanism gains better performances by margins of  0.6\%, 0.8\% and 3.5\% on Cora, Citeseer and Pubmed datasets respectively against the model without attention mechanism. As for inductive learning tasks on PPI dataset, Node2Grids applying attention mechanism also outperforms by a margin of 0.006 in \autoref{table4}. 

In addition, we further analyze the “attentions” of the grid-level attention filters. For the constructed Euclidean grid, the different grids in a channel should have different status. Due to the factor that the node degree represents the influences of nodes, we suspect the grids filled with more influential neighbors have greater weights intuitively. In  \autoref{figure6}, we present the average convergent attention filters of Cora dataset in the form of heat map, where the grids with greater value are drawn darker. Specially, the $1\times1$ convolutional kernel and $\theta_{bias} = 0$ are applied to avoid the fusion of extra nodes. It can be observed from  \autoref{figure6}(a)(b) that the grids with greater node degree have the darker color regardless of placing the sorted nodes in order or inversion. Besides, from  \autoref{figure6}(c), it can be discovered that the grids filled with central nodes have almost the same darkest color. Through the aforementioned phenomenons, we demonstrate that the attention mechanism enables specifying the weights for neighbors with different level influences implicitly. Furthermore, from the results that the more influential neighbors tend to gain more attentions, we can prove that the effectiveness of degree-based selection strategy indirectly.  

\subsection{Ablation Study}
For further exploring the effectiveness of the proposed strategies, we conduct the ablation studies to justify the contribution of each component of the proposed Node2Grids. Thus, we ablate the strategies of introducing second-order neighbors, Central-Fusion and attention mechanism. \autoref{table6} shows the quantitative comparisons of our
ablated variants, where the ablation experiments are conducted on the dataset of Cora, Citeseer and Pubmed. As we can see, each strategy boots framework independently. And the experiments obtain more positive impacts while using multiple strategies. Particularly, the intact model gains significant progress over the completely ablated framework by margins of 16.1\%, 9.4\% and 39.1\% on the dataset of Cora, Citeseer and Pubmed. This demonstrates the effectiveness of the three strategies to the proposed architecture.

\section{Conclusion}
In this paper, we present a flexible uncoupled training framework to analyze the coupled graph data, which demonstrates the great superiority in processing large-scale graph. Compared with recursive neighborhood expansion as GCN-based frameworks, Node2Grids achieves the flexibility by extracting the neighboring information at a one-off mapping step, where the mapped uncoupled data can be processed by CNN-based approaches. Through extensive experiments, we show that Node2Grids enjoys significantly lower cost in both time and memory against the GCN-based frameworks, meanwhile gains the great effectiveness on both transductive and inductive learning tasks against state-of-the-art models, which indicates that Node2Grids is more applicable than GCN-based methods in practice. Moreover, we explore the availability of the strategies applied in Node2Grids during the experiments. As a novel approach for processing graph data, we believe that it is well worth further study on account of its remarkable adaptability and scalability. In the future, it's meaningful to extend this uncoupled training framework to various graph learning tasks.

\vspace{12pt}
\color{red}

\end{document}